\RequirePackage{amsmath}
\documentclass[runningheads]{llncs}
\usepackage{graphicx}
\usepackage{times}
\usepackage{url}
\usepackage{latexsym}
\usepackage{graphicx}
\usepackage{subfigure}
\usepackage{algorithm}
\usepackage{algorithmic}
\usepackage{booktabs}
\usepackage{multirow}

\usepackage{amssymb}

\usepackage{enumerate}
\usepackage{indentfirst}
\usepackage{float}
\usepackage{algorithm}
\usepackage{algorithmic}
\usepackage{threeparttable}
\usepackage{booktabs}
\usepackage{hyperref}
\usepackage{tabularx}
\usepackage{array}

\newcommand{\tabincell}[2]{\begin{tabular}{@{}#1@{}}#2\end{tabular}}
\newcolumntype{M}[1]{>{\centering\arraybackslash}m{#1}}
\begin{document}
\title{From syntactic structure to semantic relationship: hypernym extraction from definitions by recurrent neural networks using the part of speech information}
\titlerunning{Hypernym extraction from definitions by RNN using the PoS information}
%
\author{Yixin Tan}
\author{Yixin Tan\orcidID{0000-0001-6620-3562} \and
Xiaomeng Wang\orcidID{0000-0002-7591-0127} \and
Tao Jia\orcidID{0000-0002-2337-2857}}

\authorrunning{Y. Tan et al.}

\institute{College of Computer and Information Science, Southwest University, Chongqing 400715, People’s Republic of China \\
\email{tanyixin233@gmail.com, wxm1706@swu.edu.cn, tjia@swu.edu.cn}}
\maketitle              
\begin{abstract}
The hyponym-hypernym relation is an essential element in the semantic network. Identifying the hypernym from a definition is an important task in natural language processing and semantic analysis. While a public dictionary such as WordNet works for common words, its application in domain-specific scenarios is limited. Existing tools for hypernym extraction either rely on specific semantic patterns or focus on the word representation, which all demonstrate certain limitations. Here we propose a method by combining both the syntactic structure in definitions given by the word’s part of speech, and the bidirectional gated recurrent unit network as the learning kernel. The output can be further tuned by including other features such as a word’s centrality in the hypernym co-occurrence network. The method is tested in the corpus from Wikipedia featuring definition with high regularity, and the corpus from Stack-Overflow whose definition is usually irregular. It shows enhanced performance compared with other tools in both corpora. Taken together, our work not only provides a useful tool for hypernym extraction but also gives an example of utilizing syntactic structures to learn semantic relationships \footnote{Source code and data available at \url{https://github.com/Res-Tan/Hypernym-Extraction}}.

\end{abstract}
\section{Introduction}
\noindent Hypernym, sometimes also known as hyperonym, is the term in linguistics referring to a word or a phrase whose semantic field covers that of its hyponym. The most common relationship between a hypernym and a hyponym is an ``is-a'' relationship. For example, ``red is a color'' provides the relationship between ``red'' and ``color'', where ``color'' is the hypernym of ``red''. 

The hypernym-hyponym relation is an essential element in the semantic network and corresponding tasks related to semantic network analysis \cite{hertling2017webisalod}. The hypernym graph built on a collection of hyponym-hypernym relations can enhance the accuracy of taxonomy induction \cite{navigli2011graph_lexical_taxonomies,gupta2017taxonomy}. The linkage between the hyponym and the hypernym can be used to improve the performance of link prediction and network completion in the knowledge graph or semantic network \cite{dettmers2018convolutional_completion,trouillon2017knowledge_completion}. In natural language processing (NLP), the hyponym-hypernym relation can help the named entity recognition \cite{torisawa2007exploiting_NER}, and the question-answering tasks for ``what is'' or ``is a'' \cite{saggion2004mining,cui2007soft_pattern}. The data mining, information search and retrieval can also benefit from the hyponym-hypernym relation \cite{paulheim2012unsupervised,chandramouli2008query_refinement}.

Given the role and application of the hypernym-hyponym relation, it is essential to explore an automatic method to extract such the relation between two entities, which presents an important task in knowledge-driven NLP \cite{zhang2019ernie}. Following the landmark work focusing on lexico-syntactic patterns \cite{hearst1992automatic}, several pattern-based methods are developed for hypernym extraction \cite{snow2005learning,cui2007soft_pattern}. Then the feature-based classification methods are introduced \cite{boella2013extracting_SVM,espinosa2015hypernym_grammar}, which applies machine learning tools to enhance the recall rate. Recently, distributional methods and hybrid distributional models are successfully applied to learn the embedding of words, based on which the hypernym-hyponym relation can be inferred \cite{fu2014learning_WE,shwartz2016hypernyms,shwartz2016improving}. The deep learning approach is also effective in many sequence labeling tasks including hypernym extraction \cite{li2016definition_DElstm,sun2019extracting_TwoPhase}.

While the extraction of hyponym-hypernym relation can be done in many different environments, in this work we focus on the hypernym extraction from definitions. More specifically, the definition refers to a short statement or description of a word. Take the word ``red'' as an example, whose definition on Wikipedia \footnote{\url{https://www.wikipedia.org/}} is ``Red is the color at the end of the visible spectrum of light, next to orange and opposite violet.'' The aim is to identify the word ``color'' as the hypernym of ``red'' from all the nouns in the definition. Intuitively, this task can be solved by general resources such as WordNet dictionary \cite{fellbaum1998wordnet} or Wikipedia. But given a word's different meanings in different contexts, these resources can not sufficiently complete this task. As an example, the term ``LDA'' in Wikipedia denotes ``Linear Discriminant Analysis'' in machine learning, ``Low dose allergens'' in medicine, and ``Landing distance available'' in aviation. The combination of general resources and context identification would also fail in some domain-specific applications where the general resources do not cover the special or technical terms in that area. Moreover, existing technical approaches also demonstrate certain limitations in the task of hypernym extraction from definitions, which we summarize as follows:
\begin{enumerate}[1)]
\item Hypernym and hyponym are connected in many different ways. Even the ``is a'' pattern, which is usually considered typical, has many variations such as \\``is/was/are/were + a/an/the''. It is impossible that one enumerates all different patterns. Consequently, despite high precision, the pattern selection method usually gives a low recall value.
\item The traditional feature-based classification method relies on manually selected features and the statistical machine learning models. It may work well in a class of formats, but in general, the performance can not be guaranteed once the data or the environment changes.
\item The distributional method, which relies on the similarity measure between two words to gauge the semantic relationship, is usually less precise in detecting a specific semantic relation like hypernym. Moreover, it needs a large training corpus to accurately learn the representation of words from their heterogeneous co-occurrence frequencies. In definitions, however, the appearance frequency of a word is usually low and the size of data is relatively small. The distributional method may not be directly applicable to this scenario.
\item The deep learning method, such as the recurrent neural network (RNN), can be used to process word sequences, which does not rely on particular features selected. To a great extent, it overcomes the limitation 2). However, current approaches usually take the word sequence as the input, or focus on the modification of RNN structures. Other features of the word, such as its part of speech, are not fully explored.
\end{enumerate}

To briefly illustrate the difficulty, let us consider a definition from the Stack-Overflow\footnote{\url{https://stackoverflow.com/}} with an irregular format: ``fetch-api: the fetch API is an improved replacement for XHR''. The term ``fetch-api'' is not included in any common dictionary. While the definition has the ``is an'' pattern, it does not connect to the hypernym. The definition is very short and every distinct word in this definition appears just once, which makes it difficult to accurately learn the word representation. Overall, it is challenging to find a method that would accurately identify ``API'' as the correct hypernym.

The definition of a word represents a certain type of knowledge extracted and collected from disordered data. Indeed, there are tools capable of extracting definitions from the corpora with good accuracy \cite{navigli2010learning_WCL,boella2013extracting_SVM,li2016definition_DElstm,espinosa2015hypernym_grammar,sun2019extracting_TwoPhase}. Nevertheless, tools to extract hypernym from definitions remain limited. 
To cope with this issue, we propose a recurrent network method using syntactic features. Because the definition directly points to a noun, the hyponym is already given. Therefore, the hypernym extraction is to identify the correct hypernym from all words in the definition sentence. This task can be considered as a binary classification, in which the classifier judges if a candidate noun is a hypernym or not. In order to better learn the syntactic feature, we transfer the definition sentence into the part of speech (PoS) sequence after labeling the PoS of each word by a standard tool (Stanford-NLP \cite{qi2018universal_stanford}). The syntactic structure surrounding the candidate is learned by a bidirectional gated recurrent units (GRU) based model. To further fine tune the results, we use a set of features including the centrality of the word in the hypernym co-occurrence network. We use two corpora to evaluate our method. One is Wikipedia, featuring definitions with canonical syntax structure and intensively used by previous studies. The other is from Stack-Overflow, whose definition is domain-specific and usually with the irregular format. Our method is compared with several existing ones. Overall, it outperforms all others in both corpora, which demonstrates the advantage of combing both the tool of RNN and the PoS information in the task of hypernym extraction.

This paper is organized as follows. We review related works in Section \ref{sec: related_work} and introduce details of the method in Section \ref{sec: method}. Experiments and evaluations of the proposed model are presented in Section \ref{sec: experiment}. After that, we draw a conclusion about this research in Section \ref{sec: conclusion}.

\section{Related Work}
\label{sec: related_work}
\noindent The existing methods in hypernym extraction generally fall into one of the following four categories: pattern-based method, feature-based classification method, distributional method and deep learning method. 

\subsection{Pattern-based Method}
\label{sub: pattern-based}
\noindent The pattern-based method directly uses the syntactic patterns in definitions, such as ``is-a'', ``is called'', ``is defined as'' and more. This method is commonly applied in early works due to its simplicity and intuitiveness. The majority of these approaches apply the symbolic method that depends on lexico-syntactic patterns or features \cite{hearst1992automatic}, which are manually crafted or semi-automatically learned. However, because only a small fraction of syntactic patterns can be included, these methods usually have a low recall value. In order to cover more patterns, \cite{westerhout2007extraction_usingPoS} considers PoS tags instead of simple word sequences, which raises the recall rate. To improve the generalization of the pattern-based method, \cite{cui2007soft_pattern} starts to model the pattern matching as a probabilistic process that generates token sequences. Moreover, \cite{navigli2010learning_WCL} proposes the three-step use of directed acyclic graphs, called Word-Class Lattices (WCLs), to classify definitions on Wikipedia. To better cluster definition sentences, the low-frequency words are replaced by their PoS. For a simple example, definitions that ``Red is a color'' and ``English is a language'' are in the same class that is characterized by a pattern ``noun is a noun". In this way, more patterns can be characterized to identify the hypernym. In recent years, much research pay attention to extracting hypernyms from larger data resources via the high precise of pattern-based methods. \cite{seitner2016large} extract hypernymy relations from the CommonCrawl web corpus using lexico-syntactic patterns. In order to address the low recall of pattern-based method in large data resources, \cite{shwartz2016improving,bernier2018crim} integrate distributional methods and patterns to detect hypernym relations from several existing datasets.

Nevertheless, the pure pattern-based approaches are generally inefficient, given the fact that syntactic patterns are either noisy by nature or domain-specific. It is very difficult to further improve the performance in this direction.

\subsection{Feature-based classification Method}
\label{sub: classification-based}
\noindent To overcome the issue of generalization in the pattern-based method, the feature-based classification method is introduced. \cite{navigli2010annotated} proposes a method to learn the generalized lexico-syntactic pattern and assign scores to candidate hypernyms. The scores are used to identify the true hypernym out of others. \cite{jin2013mining_CRF} uses conditional random fields to identify scientific terms and their accompanying definitions. Moreover, \cite{boella2013extracting_SVM} uses the role of syntactic dependencies as the input feature for a support vector machine (SVM) based classifier. \cite{espinosa2015hypernym_grammar} explores the features in the dependency tree analysis.

These feature-based classification approaches heavily rely on manually specified features. Patterns learned from sentences or features analyzed from the NLP tools may not fully represent the syntactic structure. In addition, the NLP tools like dependency tree analysis are often time-consuming, and error at early steps may propagate which eventually leads to inaccurate final results.

\subsection{Distributional Method}
\label{sub: distribution-based}
\noindent The distributional method is based on the Distributional Inclusion Hypothesis which suggests that a hypernym tends to have a broader context than its hyponyms \cite{kotlerman2010directional,lenci2012identifying}. If the similarity between two words can be accurately measured, then a hypernym should be associated with a similar but larger set of words than its hyponyms \cite{lenci2008distributional,lenci2012identifying,yu2015learning}, \cite{roller2014inclusive_test}tests the Distributional Inclusion Hypothesis and find that hypothesis only holds when it is applied to relevant dimensions. Because word embedding can reflect the corresponding semantic relationship, \cite{fu2014learning_WE} constructs semantic hierarchies based on the notion of word embedding. \cite{roller2016relations} uses linear classifiers to represent the target words by two vectors concatenation. \cite{chang2017distributional} introduces a simple-to-implement unsupervised method to discover hypernym via per-word non-negative vector embeddings. \cite{wang2020birre} proposes a novel representation learning framework, which generates a term pair feature vectors based on bidirectional residuals of projections, reaches a state of the art performance in general resources.

Nevertheless, the application of the distributional method relies on a very large corpus to learn the word representation. Moreover, the Distributional Inclusion Hypothesis may not be always hold. In the task discussed in this paper, because many terminologies occur infrequently and the length of a definition is usually short, it can be very inefficient to learn word representation.

\subsection{Deep Learning Method}
\label{sub: RNN-based}
\noindent The recurrent neural networks (RNN) \cite{elman1990finding_RNN} have been applied to handle many sequential prediction tasks. By taking a sentence as a sequence of tokens, RNN also works in a variety of NLP problems, such as spoken language understanding and machine translation. It is applied in hypernym extraction as well. \cite{li2016definition_DElstm} converts the task of definition extraction to sequence labeling. Using a top-N strategy (same as \cite{navigli2010learning_WCL}), the infrequently appeared words are replaced by their corresponding PoS. The sequence mixed with words and PoS elements is fed to the long short-term memory (LSTM) \cite{hochreiter1997long_lstm} RNN to predict the definition.
More recently, \cite{sun2019extracting_TwoPhase} proposes a two-phase neural network model with yields an enhanced performance compared with \cite{li2016definition_DElstm}. The first phase is constructed by a bi-directional LSTM to learn the sequence information. Then a CRF and a logistic regression are used to refine the classification results. 
Both of the two works focus on words. Although \cite{li2016definition_DElstm} considers the PoS information, the purpose is only to reduce the total number of words by grouping less frequent words together according to their PoS property. While they demonstrate improved performance compared with other methods, they are only tested in Wikipedia corpus, where the definition usually has a very regular format. The performance on other irregular definitions remains unknown.

\section{Method}
\label{sec: method}

\begin{figure}
\centering
\includegraphics[width=\linewidth]{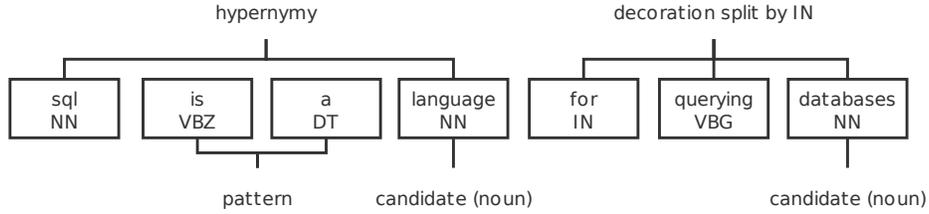}
\caption{An example of a hypernym-hyponym pair in a definition: ``sql is a language for querying databases''. The definition is split into units (words and the corresponding PoS) for analysis. The word ``language'' and ``databases'' are two hypernym candidates. The PoS elements surround ``language'' and ``databases'' are different. Our model learns such features and identifies ``language'' as the hypernym of ``sql''.}
\label{fig: example}
\end{figure}

\begin{figure}
\centering
\includegraphics[width=\linewidth]{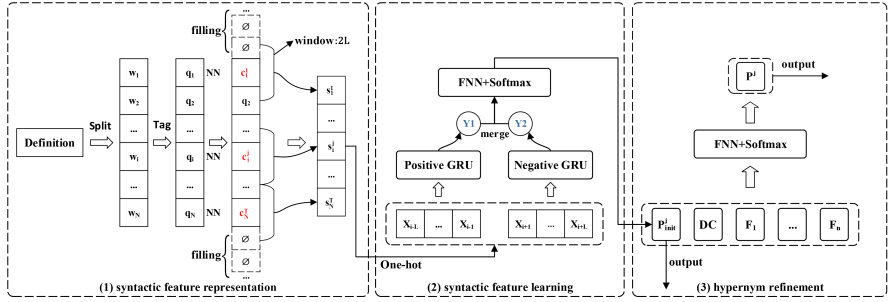}
\caption{The architecture of our model includes three phases: (1) syntactic feature representation (2) syntactic feature learning and (3) hypernym refinement.}
\label{fig: model}
\end{figure}

\noindent In our approach, a definition sentence is split into words. The words are further labeled according to their grammatical properties, which form a PoS sequence representing the syntactic structure of the definition. The nouns are selected as hypernym candidates which need to be classified. An illustration of this procedure is shown in Figure \ref{fig: example}. We particularly focus on the syntactic structure surrounding a noun. This feature is learned from the training set that helps the hypernym recognition in the testing set. Our model contains three phases (Figure \ref{fig: model}): syntactic feature representation, syntactic feature learning, and hypernym identification refinement.

\subsection{Syntactic Feature Representation}
\noindent In the first phase of hypernym extraction, a definition sentence is converted into a context segment sequence which captures syntactic features of the definition. The context segment sequence is used as the input of the RNN at the second phase. 

A definition sentence can be considered as a word sequence of $N$ elements $W=[w_1,..., w_i,..., w_N]$, which further gives a PoS sequence $Q=[q_1,..., q_i,..., q_N]$. Assume that there are $T$ nouns in the definition which are the hypernym candidates. These $T$ nouns can be recorded as $C = \{ c^j_i \}$, where $i$ is the position of the noun in the word sequence and $j$ is its order in the $T$ nouns. We use a window to extract the local syntactic feature around a noun from the PoS sequence $Q$, yielding $T$ context segments as
\begin{equation}
\label{eq: context_segment}
s^j_{i} = [q_{i-L}, ..., q_{i-1},q_{i+1}, ..., q_{i+L}],
\end{equation} 
where $L$ is the window size which also determines the length of each context segment. To make each context segment equal length, we extend the sequence $Q$ by adding the null element on its two ends when needed, i.e. $q_i =\varnothing$ for $i < 1$ and $i > N$.

Because the number of PoS types is limited and small, we can represent each $q_i$ as a one-hot vector $X_{i}$, where the corresponding PoS type has the value 1 and others are with value 0. More specifically, in this work, we consider 15 PoS types and one null element $\varnothing$. Consequently, each $q_i$ is represented by a 16-dimensional vector $X_{i}$ and $s^j_{i}$ is represented by equation \ref{eq: context_matrix}, which is a 16 by $2L$ matrix. 
\begin{equation}
\label{eq: context_matrix}
s^j_{i} = [X_{i-L}, ..., X_{i-1},X_{i+1}, ..., X_{i+L}],
\end{equation}

\subsection{Syntactic Feature Learning}
\noindent We use the RNN to learn the local syntactic features. Because the original RNN model cannot effectively use the long sequential information due to the vanishing gradient problem \cite{Hochreiter01gradientflow}, the long short-term memory (LSTM) architecture is proposed to solve this issue. In our input, a context segment $s^j_i$ can be divided into two parts: the pre-sequence $[X_{i-L}, ..., X_{i-1}]$ and the post-sequence $[X_{i+1}, ..., X_{i+L}]$. Naturally, we adopt the gated recurrent unit (GRU) \cite{chung2014empirical_GRU} architecture, which is a variant of LSTM, but simpler, and faster in training than LSTM. We use a bi-directional structure (Figure \ref{fig: model}(2)) containing a positive GRU and a negative GRU to learn the pre- and post-syntactic features separately from the above two sequences. The intermediate results $Y_1$ and $Y_2$ obtained through the two GRU modules are merged into $Y=[Y_1;Y_2]$ and fed into a feedforward neural network. The softmax layer outputs the probability $P^j_{init}$ that $c^j_{i}$ is the hypernym. $P^j_{init}$ can be expressed as
\begin{equation}
\label{eq: candidate}
  P^j_{init} = p(c^j_i|s^j_i) = p(c^j_i|X_{i-L},...,X_{i-1},X_{i+1},...,X_{i+L})
\end{equation}

\subsection{Hypernym Refinement}
\noindent The initial probability $P^j_{init}$ obtained through the above steps can be used directly to identify the hypernym. Nevertheless, some other features of the words can be used to improve accuracy. The $P^j_{init}$ and the selected features are fed into another feedforward neural network to compute the final probability $P^j$, which is presumably more optimal. The candidate with the maximum probability is selected as the hypernym of the target definition.

\begin{figure}
\centering
\includegraphics[width=\linewidth]{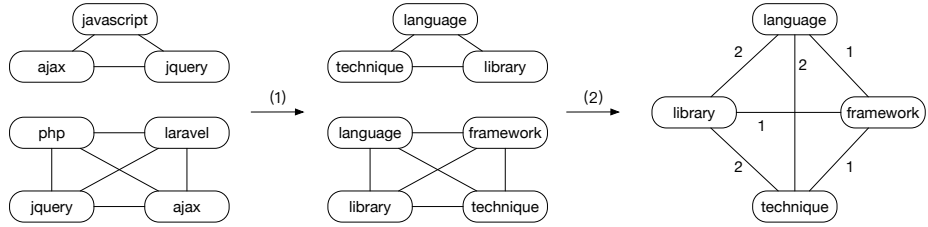}
\caption{A simple example of the hypernym graph construction process. (1): terms of co-occurrence are replaced by their corresponding hypernyms from the training set. (2): hypernym co-occurrence network is built based on the co-occurrence of the hypernym.}
\label{fig: hypernym_graph}
\end{figure}

Features that can be included in this phase include a word's position in the sentence, whether it is capitalized, the frequency of usage, and so on. We encode these as a refinement feature vector $[F_1,F_2,...,F_n]$. Besides these commonly known features, we also consider the degree centrality (DC) of a candidate in the hypernym co-occurrence network, following the intuition that a concept with higher centrality in a semantic network is more likely to be a hypernym. In the folksonomy, such as Stack-Overflow and Twitter, an item may be tagged by multiple labels \cite{wang2020measuring}. A scientific paper may also be labeled with multiple keywords or other tags \cite{jia2017quantifying}. The fact that multiple entities simultaneously occur together tells some hidden relationship between them. To make use of this feature, we first extract the co-occurrence of hyponyms from the data, where multiple hyponyms are used as notations of a question or a statement. Using the hyponym-hypernym relationship in the training set, we further obtain the co-occurrence of the hypernym, based on which the hypernym co-occurrence network is built. Figure \ref{fig: hypernym_graph} gives an example of the hypernym co-occurrence network construction. The feature DC, which counts how many neighbors a hypernym has, can help identify hypernyms in several tricky cases. For example, the definition ``fetch-api: the fetch API is an improved replacement for XHR, ...'', $P_{init}$ would predict ``replacement'' as the hypernym. The real hypernym ``API'' can only be revealed after taking the DC feature into consideration.

\section{Experiment}
\label{sec: experiment}
\noindent We test and evaluate our method with both Wikipedia and Stack-Overflow data sets. Before the experiment, some details about data are introduced to explain the basis of feature selection. Then, we compare the performance of our method with other existing ones. Finally, we perform extended tests to confirm the advantage of using syntactic features and the RNN in hypernym extraction.

\subsection{Dataset}
\label{sub: dataset}

\begin{table}[htbp]
\centering
\caption{Details of annotation datasets from Wikipeida and Stack-Overflow.}
\setlength{\tabcolsep}{4mm}{
\begin{tabularx}{\linewidth}{llllll}
  \toprule
  Dataset & Definitons & \tabincell{l}{Invalid-\\definitions} & \tabincell{l}{Total\\words} & \tabincell{l}{Total\\sentences} & \tabincell{l}{Average\\length} \\
  \midrule
  Wikipedia & 1871 & 2847 & 21843 & 4718 & 12.05 \\
  Stack-Overflow & 3750 & 1036 & 9921 & 4786 & 14.29 \\
  \bottomrule
\end{tabularx}}
\label{tab: datasets_details}
\end{table}

\noindent Two corpora are selected to train and test our method. One is the public Wikipedia corpus \cite{navigli2010annotated} and the other is the corpus from Stack-Overflow. The definition syntax in Wikipedia is very standardized. Hence the Wikipedia corpus is used in most existing works. However, besides common concepts, domain-specific concepts or terms are emerging from different fields. One typical example is computer science. In the online community Stack-Overflow, massive technical terms are discussed and organized, providing a rich body of definition corpus. In this work, we collect about 36,000 definitions from Stack-Overflow. The details of annotation datasets are shown in Table \ref{tab: datasets_details}.

\begin{table}[htbp]
\centering
\caption{15 PoS and their corresponding abbreviations in our experiment.}
\setlength{\tabcolsep}{7mm}{
\begin{tabularx}{\linewidth}{ll}
  \toprule
  Abbreviation & PoS \\
  \midrule
  DT & Determiner \\
  EX & Existential \textit{there} \\
  IN & Preposition or subordinating conjunction \\
  NN & Noun (singular or plural), Proper Noun (singular or plural) \\
  TO & to \\
  VB & Verb, base form \\
  VBD & Verb, past tense \\
  VBG & Verb, gerund or present participle \\
  VBN & Verb, past participle \\
  VBP & Verb, non-3rd person singular present \\
  VBZ & Verb, 3rd person singular present \\
  WDT & Wh-determiner \\
  WP & Wh-pronoun \\
  WP\$ & Possessive wh-pronoun \\
  WRB & Wh-adverb \\
  \bottomrule
\end{tabularx}}
\label{tab: pos_usage}
\end{table}

Some data pre-processing is performed. First, we use the definition extraction method \cite{sun2019extracting_TwoPhase} to filter out invalid definitions. Second, we remove words in the parentheses because they are usually used for explanations and no likely to contain the hypernym. For example, the sentence ``Javascript (not be confused with Java) is a programming language ...'' is simplified to ``Javascript is a programming language ...''. In addition, we remove some PoS such as adjectives and adverbs after PoS Tagging, which would not affect the meaning of a text. The 15 PoS types used in our methods are shown in Table \ref{tab: pos_usage}.

\subsection{PoS Position Comparison}
\noindent To demonstrate that the syntactic structure captured by the PoS elements is a suitable feature for hypernym identification, we show the probability that a PoS element appears around a hypernym and a non-hypernym (Table~\ref{tab: position_illustration}). For simplicity, we only consider the closest word before and after the hypernym and the non-hypernym (equivalently window size $L=1$ in our model). For non-hypernyms, except for WDT and DT, a PoS element appears on either side with roughly the same probability. In contrast, the appearance of the PoS element around the hypernym is very polarized. For example, for more than 99\% of the time, a preposition appears after the hypernym. The clear difference in the syntactic structure surrounding the hypernym and non-hypernym provides a good basis for the classification task.

\begin{table}[htbp]
\centering
\caption{The probability that a PoS element appears before ($P_1$) and after ($P_2$) a target. The probability is conditioned on the appearance of the PoS element hence $P_1 + P_2 = 1$. $N$ represents the cases that the target is not a hypernym and $H$ represents that the target is a hypernym. }
\setlength{\tabcolsep}{7mm}{
\begin{tabularx}{\linewidth}{lcccc}
  \toprule
  PoS & $P_1(N)$ & $P_2(N)$ & $P_1(H)$ & $P_2(H)$ \\
  \midrule
  WDT & 0.065 & 0.935 & 0 & 1 \\
  IN & 0.571 & 0.429 & 0.008 & 0.992 \\
  TO & 0.540 & 0.460 & 0.028 & 0.972 \\
  VBP & 0.539 & 0.461 & 0.033 & 0.967 \\
  VBZ & 0.404 & 0.596 & 0.044 & 0.956 \\
  VBN & 0.385 & 0.614 & 0.071 & 0.929 \\
  VBG & 0.647 & 0.353 & 0.428 & 0.572 \\
  NN & 0.416 & 0.584 & 0.963 & 0.037 \\
  DT & 0.933 & 0.067 & 0.970 & 0.030 \\
  \bottomrule
\end{tabularx}}
\label{tab: position_illustration}
\label{sub: pos_comparison}
\end{table}

\subsection{Method Comparison and Evaluation}
\label{sub: method_comeparison}
\subsubsection{Baseline Methods.} To illustrate that the PoS based feature is more effective than the word-based feature, we separately take the one-hot code of PoS and the embedding of the word as input. The two models with different inputs are denoted by $\text{Model}_\text{PoS}$ and $\text{Model}_\text{Word}$. We also consider other existing methods for comparison, including \textbf{(1) WCLs}: An algorithm that learns a generalization of word-class lattices for modeling textual definitions and hypernym \cite{navigli2010learning_WCL}. \textbf{(2) Dependencies}: A method that only uses syntactic dependencies features extracted from a syntactic parser to fed into the classifier and extract definitions and hypernyms \cite{espinosa2015hypernym_grammar}. \textbf{(3) Grammar}: A feature engineering model for hypernym extraction, using 8 handcrafted features which contain linguistic features, definitional features and graph-based features \cite{espinosa2015hypernym_grammar}. \textbf{(4) Two-Phase}: A deep learning model for sequence labeling hypernym extraction based on bidirectional LSTM and CRF \cite{sun2019extracting_TwoPhase}.

\begin{table}[htbp]
\centering
\caption{Hypernym Extraction in Wikipedia corpus and Stack-Overflow corpus: the best results are shown in \textbf{black bold} and $\text{Model}_\text{Word}$ is used as comparison.}
\begin{threeparttable}
  \setlength{\tabcolsep}{6mm}{
  \begin{tabularx}{\linewidth}{clccc}
  \toprule
  Dataset & Method & P \% & R \% & F1 \% \\
  \midrule
  \multirow{6}{*}{Wikipedia} & WCLs\cite{navigli2010learning_WCL} & 78.6 & 60.7 & 68.6 \\
  & Dependencies \cite{boella2013extracting_SVM} & 83.1 & 68.6 & 75.2 \\
  & Grammar \cite{espinosa2015hypernym_grammar} & 84.0 & 76.1 & 79.9 \\
  & Two-Phase \cite{sun2019extracting_TwoPhase} & 83.8 & 83.4 & 83.5 \\
  & $\text{Model}_\text{Word}$ & 82.1 & 76.8 & 79.4 \\
  & $\text{Model}_\text{PoS}$ & \textbf{94.4} & \textbf{88.3} & \textbf{91.3} \\
  \midrule
  \multirow{6}{*}{Stack-Overflow} & WCLs\cite{navigli2010learning_WCL} & 75.2 & 58.6 & 65.9 \\
  & Dependencies \cite{boella2013extracting_SVM} & 81.7 & 66.2 & 73.1 \\
  & Grammar \cite{espinosa2015hypernym_grammar} & 82.8 & 71.4 & 76.7 \\
  & Two-Phase \cite{sun2019extracting_TwoPhase} & 86.3 & 78.4 & 82.2 \\
  & $\text{Model}_\text{Word}$ & 76.1 & 72.9 & 74.5 \\
  & $\text{Model}_\text{PoS}$ & \textbf{94.7} & \textbf{90.2} & \textbf{92.4} \\
  \bottomrule
  \end{tabularx}}
\end{threeparttable}
\label{tab: method_comparison}
\end{table}

\subsubsection{Experimental Settings.} (1) We use 80\% of the total sample as the training set and another 20\% as the testing set. (2) The performance of a method is measured by precision (P), recall (R), and F1-Score (F1) metric. (3) Extra-features for refinement including a word's position, capitalized, usage frequency, and degree centrality. (4) In $\text{Model}_\text{Word}$, we use the embedding layer to convert each word into a vector representation by looking up the embedding matrix $W^{\text{word}} \in \mathbb{R}^{d^w\left|V\right|}$, where $V$ is a fixed-sized vocabulary, and $d^w$ is the 100-dimensional embedding size. The matrix $W^{\text{word}}$ is a parameter to be learned. We transform a word $w_i$ into its word embedding $e_i$ by using the matrix-vector product:
\begin{equation}
e_i=W^{\text{word}}v^i,
\end{equation}
where $v_i$ is a vector of size $\left|V\right|$ which has value 1 at index $e_i$ and 0 in all other positions. (5) To prevent neural networks from over fitting, a dropout layer \cite{srivastava2014dropout} is used. (6) The objective formulation is defined by Cross-Entropy, and the root mean square prop (RMSProp) \cite{tieleman2012lecture_rmsp} algorithm is used to train our model.

\subsubsection{Empirical Results.} The results (Table~\ref{tab: method_comparison}) show that the proposed method outperforms all existing ones. The different performance between $\text{Model}_\text{PoS}$ and $\text{Model}_\text{Word}$ confirms the advantage of using PoS feature in the hypernym extraction over the use of word embedding. It is noteworthy that the accuracy in PoS tagging would significantly affect the final outcome, given the role of PoS in our method. As an example, depending on the context, the word ``control'' can either be a verb or a noun. Therefore, for the definition ``gridview: a control for displaying and manipulating data from ...'', incorrectly tagging ``control'' as a verb will yield incorrect hypernym. For simplicity, the task of PoS tagging in our work is carried out by the Stanford-NLP tool. But its accuracy still has the potential for further improvement, which can eventually enhance the final performance of our method.
 
\subsubsection{Hyper-parameters Sensitivity.} 
\begin{figure}
\centering
\includegraphics[width=\linewidth]{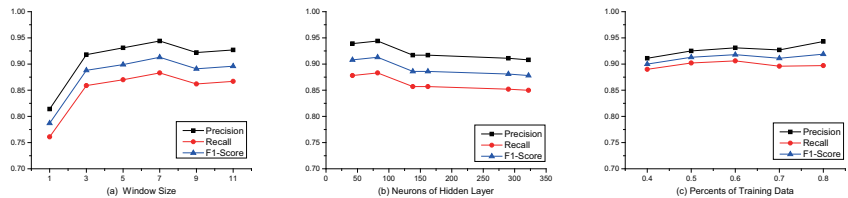}
\caption{The model performance (the Precision, Recall and F1-Score in the y-axis) with varying window sizes (a), neuron number in the hidden layer (b) and the ratio of training samples (c). }
\label{fig: params}
\end{figure}

We show the Precision, Recall and F1-Score of our model with different hyper-parameters to analyze the model's sensitivity (Figure \ref{fig: params}). In general, the choice of hyper-parameters does not significantly affect the performance of our model.

\subsection{Word Feature and Learning Kernel Ablation}
\label{sub: ablation}
\subsubsection{Hybrid Representation Strategy for Word Feature Ablation.} The fact that the\\ $\text{Model}_\text{PoS}$ outperforms the $\text{Model}_\text{Word}$ confirms the advantage of using PoS as the input feature. This, however, gives rise to another question: could the performance improve if the model combines both the PoS feature and word embedding? Indeed, the hybrid representation strategy was successfully applied in previous studies \cite{navigli2010learning_WCL,li2016definition_DElstm} to reach improved extraction results. For this reason, we analyze the performance of the hybrid strategy. For a definition sentence $W=[w_1,w_2,...,w_N]$, we convert the word $w_i$ into token $t_i$ as follows:
\begin{equation}
\label{eq: top_n}
t_i = \left\{
\begin{aligned}
& w_i & & {w_i\in W_{top}} \\
& PoS(w_i) & & {w_i\notin W_{top}}
\end{aligned} \right.
\end{equation}
where $W_{top}$ is a set of top-K words of appearance. In this way, a word $w_i$ is left unchanged if it occurs frequently in the training corpus, or it is converted into its PoS. Eventually, we obtain a generalized definition $W'=[t_1,t_2,...,t_N]$ with a mixture of words and PoS terms. 

\begin{table}[htbp]
\centering
\caption{The performance of our model after using the TOP-K strategy. In this table, $K$ represents the hyper-parameter of TOP-K strategy, W represents the Wikipedia corpus and S represents the Stack-Overflow corpus. The best results are shown in \textbf{black bold}.}
\begin{threeparttable}
\setlength{\tabcolsep}{7mm}{
\begin{tabularx}{\linewidth}{clcc}
  \toprule
  Representation & $K$ & W (F1\%) & S (F1\%) \\
  \midrule
  \multirow{9}{*}{Word Embeddings} & 25 & 88.3 & 89.6 \\
  & 50 & 88.6 & 89.5 \\
  & 100 & 89.0 & 91.0 \\
  & 200 & 89.0 & 88.8 \\
  & 400 & 89.7 & 89.1 \\
  & 800 & 90.2 & 85.8 \\
  & 2000 & 85.6 & 78.8 \\
  & 4000 & 81.4 & 76.3 \\
  & 8000 & 80.5 & 75.7 \\
  \midrule
  \multirow{5}{*}{One-hot} & 10 & 82.7 & 83.8 \\
  & 20 & 77.8 & 80.1 \\
  & 30 & 72.4 & 77.9 \\
  & 40 & 67.7 & 74.8 \\
  & 50 & 61.9 & 69.3 \\
  \midrule
  \multirow{1}{*}{$\text{Model}_\text{PoS}$} &  & \textbf{91.2} & \textbf{92.4} \\
  \bottomrule
\end{tabularx}}
\end{threeparttable}
\label{tab: top_n}
\end{table}

The $W'$ is used to replace the PoS sequence $Q$ in our method (Fig. \ref{fig: model}) which further gives the context segment $s^j_{i}$. We consider two strategies to convert the token $t_i$ into a high dimensional vector. One is to use the embedding layer to convert each term into a vector with dimension 100. The other is to use the one-hot vector to convert a top-K word into a vector with dimension $K+16$. The $s^j_{i}$ is then fed into the same GRU kernel as that in our model. The results are shown in Table \ref{tab: top_n}. Overall, word embedding is more suitable for this mixed feature representation. The performance varies on the choice of top-K values and the best parameters differ in different data sets. Nevertheless, the best performance of the hybrid strategy is not as good as our original method, which further confirms the advantage of directly using only PoS information.

\begin{table}[H]
\centering
\caption{The performances of hypernym extraction methods, which contain traditional classifiers using PoS distributional features and deep learning models using word and PoS representation. The best results are shown in \textbf{black bold}.}
\begin{threeparttable}
  \setlength{\tabcolsep}{5.9mm}{
  \begin{tabularx}{\linewidth}{clccc}
  \toprule
  Dataset & Method & P \% & R \% & F1 \% \\
  \midrule
  \multirow{13}{*}{Wikipedia} & Naive Bayes & 85.8 & 81.7 & 83.7 \\
  & LDA & 87.4 & 83.3 & 85.3 \\
  & Softmax Regression & 88.4 & 84.1 & 86.2 \\
  & SVM & 87.3 & 83.2 & 85.2 \\
  & Decision Tree & 83.1 & 79.2 & 81.1 \\
  & Random Forest & 87.9 & 83.8 & 85.8 \\
  & CRF & 88.9 & 77.0 & 82.5 \\
  \cmidrule{2-5}
  & $\text{Model}_\text{Word}$ & 82.1 & 76.8 & 79.4 \\
  & $\text{Transformer}_\text{Word}$ & 86.6 & 81.9 & 84.2 \\
  & $\text{Bert}_\text{Word}$ & 87.3 & 83.6 & 85.4 \\
  & $\text{Model}_\text{PoS}$ & 94.4 & 88.3 & 91.3 \\
  & $\text{Transformer}_\text{PoS}$ & 94.8 & 88.7 & 91.6 \\
  & $\text{Bert}_\text{PoS}$ & \textbf{95.2} & \textbf{89.1} & \textbf{92.0} \\
  \midrule
  \multirow{13}{*}{Stack-Overflow} & Naive Bayes & 84.8 & 78.4 & 81.5 \\
  & LDA & 86.0 & 81.9 & 83.9 \\
  & Softmax Regression & 87.2 & 82.3 & 84.7 \\
  & SVM & 87.7 & 83.6 & 85.6 \\
  & Decision Tree & 83.2 & 78.2 & 80.6 \\
  & Random Forest & 88.4 & 83.7 & 86.0 \\
  & CRF & 84.1 & 80.6 & 82.3 \\
  \cmidrule{2-5}
  & $\text{Model}_\text{Word}$ & 76.1 & 72.9 & 74.5 \\
  & $\text{Transformer}_\text{Word}$ & 80.6 & 74.3 & 77.3 \\
  & $\text{Bert}_\text{Word}$ & 76.1 & 71.9 & 74.8 \\
  & $\text{Model}_\text{PoS}$ & 94.7 & 90.2 & 92.4 \\
  & $\text{Transformer}_\text{PoS}$ & 95.1 & 90.6 & 92.8 \\
  & $\text{Bert}_\text{PoS}$ & \textbf{95.5} & \textbf{91.0} & \textbf{93.2} \\
  \bottomrule
  \end{tabularx}}
\end{threeparttable}
\label{tab: classifier_comparison}
\end{table}

\subsubsection{Learning Kernel Ablation.} While the RNN model adequately solves the problem, it is not the most up-to-date tool in sequence labeling. The recent pre-training language models such as Bert \cite{devlin2018bert}, which is based on the Transformer structure \cite{vaswani2017attention}, has led to significant performance gains in many NLP applications \cite{liu2019roberta}. Hence, it is of interest to analyze to what extend the final performance can be improved if the learning kernel is replaced by Transformer or by Bert. For this reason, we perform a learning kernel ablation experiment by applying the Transformer encoder and Bert encoder kernels in our model. We use the same input of word embedding and PoS feature as these used in $\text{Model}_\text{Word}$ and $\text{Model}_\text{PoS}$. Correspondingly, the results are recorded as $\text{Transformer}_\text{Word}$, $\text{Transformer}_\text{PoS}$, $\text{Bert}_\text{Word}$ and $\text{Bert}_\text{PoS}$.

In addition, to bring some insights on extent that our results benefit from the deep learning kernels, we apply some traditional classifiers and compare the results with deep learning kernels. For the traditional classifiers, we focus on the PoS feature captured by the context segment $s_i^j$ which is extracted from the PoS sequence $Q=[q_1,..., q_i,..., q_N]$. In our RNN based method, each PoS element $q_i$ is converted to a one-hot vector. Consequently, $s_i^j$ becomes a 16 by $2L$ matrix where the number 16 corresponds to the 15 PoS elements and a and a null element $\varnothing$. To make the input compatible with traditional classifiers, we consider a slightly different representation of $s_i^j$. We use an integer $I_q$ from 1 to 16 to represent each of the 16 possible values of $q$. To distinguish the complementary relationship that an element is before the noun and after the noun, we represent the pre-sequence $[q_{i-L}, ..., q_{i-1}]$ as $[I_{q_{i-L}}, ..., I_{q_{i-1}}]$ and the post-sequence $[q_{i+1}, ..., q_{i+L}]$ as $[33-I_{q_{i+1}}, ..., 33-I_{q_{i+L}}]$. In addition, we insert the same set of features $[F_1,...,F_n]$ used in the refinement phase to the end of the sequence $s_i^j$. In this way, the $s_i^j$ is converted into a one-dimensional vector as $[I_{q_{i-L}}, ..., I_{q_{i-1}}, 33-I_{q_{i+1}} ..., 33-I_{q_{i+L}},DC, F_1, ..., F_n]$.

The results by different deep learning kernels and traditional classifiers are shown in Table \ref{tab: classifier_comparison}. When fixing the PoS feature as the input, the use of RNN at least improves the F1 score by about 6 percentiles compared to traditional classifiers. The improvement by Transformer and Bert over RNN is relatively marginal, which is roughly 1 percentile. It is somewhat expected that Transformer and Bert will give better results, as these two kernels are more sophisticated. The magnitude of the improvement, however, implies that RNN might be a better balance between the performance and the computational complicity.
Furthermore, the comparison between results by different types of input clearly demonstrates the advantage of using the PoS feature. Indeed, random forest, a very simple classifier but with PoS feature as the input, can easily outperform the deep learning kernels with the word embedding input ($\text{Model}_\text{Word}$, $\text{Transformer}_\text{Word}$ and $\text{Bert}_\text{Word}$) in both data sets. While the word representation is almost the by-default approach in related studies, the results presented in Table \ref{tab: classifier_comparison} shows that using the right choice of input can sometimes be more efficient than optimizing the architecture of the learning kernel.

\section{Conclusion and Future Work}
\label{sec: conclusion}
\noindent The hyponym-hypernym relationship plays an important role in many NLP tasks. Despite intensive studies on this topic, tools that can accurately extract hypernym from a definition is limited. The definition, representing a special type of summarized knowledge, is commonly observed, not only because some corpora such as Wikipedia or GitHub directly give the definition of a term, but also because there are tools capable of extracting definitions with good accuracy. Hence, it is useful to develop a capable tool for this task. Here we construct a bidirectional GRU model for patterns learning. We use the PoS tags of words surrounding the hypernym as the feature. Our model outperforms existing methods in both the general corpus (Wikipedia) and the domain-specific corpus (StackOverflow). It also demonstrates a good balance between the performance and complexity, if compared with the kernels by Transformer or Bert. More importantly, by the feature and kernel ablation, we show that the PoS feature is indeed the key element that guarantees the final performance. 

The application of the tool we proposed in Stack-Overflow would help us understand the evolution of technology, group users for social network study, and build the semantic network in the domain of computer science. The performance of the tool is limited by the accuracy of PoS tagging. Hence, it would be useful to try or develop other methods other than the Stanford-NLP tool. The use of PoS feature may also have potential in other text sequence labeling tasks, which may have advantages over the word embedding. All these problems will be addressed in future studies.

\section{Acknowledgments}
This work is supported by the Fundamental Research Funds for the Central Universities (No. XDJK2017C026).

%
%
\bibliographystyle{splncs04}
%

\end{document}